# Instance-based entropy fuzzy support vector machine for imbalanced data

Poongjin Cho[†], Minhyuk Lee[†], Woojin Chang[†,*]

[†] Department of Industrial Engineering, Seoul National University, Seoul 151-742, Republic of Korea

* Corresponding Author

| | |
|---|---|
| Postal Address: | 1 Gwanak-ro, Gwanak-gu, Seoul, Republic of Korea |
| Email Address: | nadapj@snu.ac.kr (P. Cho) |
| | minhyuk.lee@snu.ac.kr (M. Lee) |
| | changw@snu.ac.kr (W. Chang \| **Corresponding**) |
| Telephone numbers: | +82) 2-880-8335 |

**Abstract**

Imbalanced classification has been a major challenge for machine learning because many standard classifiers mainly focus on balanced datasets and tend to have biased results towards the majority class. We modify entropy fuzzy support vector machine (EFSVM) and introduce instance-based entropy fuzzy support vector machine (IEFSVM). Both EFSVM and IEFSVM use the entropy information of $k$-nearest neighbors to determine the fuzzy membership value for each sample which prioritizes the importance of each sample. IEFSVM considers the diversity of entropy patterns for each sample when increasing the size of neighbors, $k$, while EFSVM uses single entropy information of the fixed size of neighbors for all samples. By varying $k$, we can reflect the component change of sample's neighbors from near to far distance in the determination of fuzzy value membership. Numerical experiments on 35 public and 12 real-world imbalanced datasets are performed to validate IEFSVM and area under the receiver operating characteristic curve (AUC) is used to compare its performance with other SVMs and machine learning methods. IEFSVM shows a much higher AUC value for datasets with high imbalance ratio, implying that IEFSVM is effective in dealing with the class imbalance problem.

# 1. Introduction

The classification problem for imbalanced datasets is considered a major challenge in machine learning [1]. The first reason arises from the problem of ignoring minority data. Handling imbalanced data through a standard classification model is biased to the majority data [2]. This is because the standard classification model assumes that the training data has a balanced class distribution [3]. In addition, the performance measure of the standard classification is provided with a better score in the majority class because it is adjusted by the standard accuracy rate. In imbalanced problems, however, the importance of the minority class is generally greater than the majority class [4]. In addition, minority data is difficult to obtain and, if not properly classified, it renders significant costs. That is, it should not set the same misclassification cost between the classes. Second, there is a problem that can be intuitively considered if it is converted to imbalanced data. For example, in the case of a multi-class problem, the number of elements in a class cannot be the same, and there are some minority classes. In this case, it is better to look at several imbalanced binary problems than to recognize them as multi-class problems [5]. Therefore, imbalanced classification should be handled differently from standard classification, and since there exists innumerable applicability to other fields, many solutions have been proposed to solve this problem.

It is well accepted to use fuzzy support vector machine (FSVM) to handle the imbalanced problem, because weights of FSVM can be given differently for each sample [6]. There are many studies on how to set the fuzzy membership of FSVM by giving low weights to the majority sample and high weights to the minority sample, so that they can make a difference in importance when making models. Batuwita and Palade [7] first developed the fuzzy penalty with the distance from the hyperplane and from the center of the class.

From another perspective, entropy is known to have a descriptive power of data and to quantify the degree of information's certainty [8]. Recently, Boonchuay et al. [9] developed minority entropy which uses information from the minority class with decision tree induction to handle the class imbalance problem. In the field of data mining, studies combining entropy and nearest neighbors have been carried out. Kaleli [10] introduced the entropy-based neighbor selection approach which collects the most similar neighbors with the smallest entropy difference. Also, there is an attempt to obtain the entropy value by examining the class ratio of the nearest neighbors, for example, Chen et al. [11] first introduced how to approximate uncertainty through entropy in a neighborhood system. Thus, neighborhood entropy can be a measure of the degree to which the class is certain, so that it can be used to formulate the membership of the FSVM.

In this paper, we propose an instance-based entropy fuzzy support vector machine (IEFSVM), an improvement to the entropy-based fuzzy support vector machine (EFSVM) [12]. EFSVM examines the class ratio of nearest neighbors for each sample, and obtains entropy value based on that. This is called the nearest neighbors entropy, and fuzzy membership is set by this value. The performance of the model is compared with the existing classification method through area under the receiver operating characteristic curve (AUC). The parameter of this EFSVM model is the number of nearest neighbors, $k$, and EFSVM uses a uniform $k$ for all samples. It is clear that most of the existing literatures [13-17] assigned weights to the instances, focusing on setting appropriate function of $k$, instead of considering the change of class ratio according to neighborhood size. However, using a unified $k$ does not work for some samples and may lead to misclassification [18]. If $k$ is too small, the useful information may be insufficient, whereas a large value of $k$ results in outliers to be included in the $k$ nearest neighbors [19]. Thus, in order to design a classifier with less sensitive to the neighborhood size, several literatures [18, 20-22] combine information by neighborhood size. Instead of tuning $k$ as a fixed value, it makes sense to combine the information in response to various neighborhood size to evaluate fuzzy membership, and this motivates us to develop an appropriate combination of the nearest neighbors entropy.

The proposed IEFSVM better reflects nearest neighbors' information considering the component change of nearest neighbors' classes in response to increasing the neighborhood size. We develop certain combination technique of nearest neighbors entropy by neighborhood size, which enables us to reflect all information on fuzzy membership. This can be achieved by exploring the characteristics of the nearest neighbors entropy. We analyze the pattern of all nearest neighbors entropy that can be generated by changing the neighborhood size. For each instance, the entropy combination of several nearest neighbors is performed, and the pattern of the combination is represented by graphical analysis. With the analysis, we provide rational reasoning to the combination formula for fuzzy membership. Unlike [12], there is no need to tune the neighborhood size to minimize the validation error of the classifier. We undertake extensive experiments on public as well as real datasets to evaluate the performance of the proposed methods through AUC. We also compare the performance of proposed methods with other state-of-the-art algorithms.

The rest of this paper is organized as follows. Section 2 presents a brief review of the main related works. In Section 3, we describe the proposed approach (IEFSVM) and the component of IEFSVM for comparison. Empirical study, results, and application to real-world datasets are reported in Section 4. The conclusion is provided in the Section 5.

## 2. Related works

In many real-world classification applications, there are numerous class imbalance problems, which means the number of samples greatly varies between different classes [23]. Imbalance problems are an important issue in many areas of analysis such as biology [24, 25], ecology [26, 27], finance [28, 29], marketing [30], medicine [31, 32], telecommunication [33] and the web [34]. It is widely known to use FSVM to handle the imbalanced problem, so that there are many literatures on how to assign the fuzzy membership of FSVM. Jiang et al. [35] proposed a fuzzy membership calculated in the feature space not in the input space, and can be represented by the kernel function. Dai [36] developed fuzzy penalty according to distance function and decaying function, and combined total margin algorithm with SVM based on the penalty. Lee et al. [37] adjusted weights in AdaBoost with a weak learner of weighted support vector machine. Imbalance Ratio (IR) is defined as the ratio of the number of minority class samples and the majority ones, *i.e.* $IR = n_{neg}/n_{pos}$ where $n_{neg}$ and $n_{pos}$ are the number of negative (minority) and positive (majority) samples, respectively. If the data is imbalanced, IR has a value greater than one. Hwang et al. [38] used this IR value in determining the fuzzy membership.

Alternatively, the nearest neighbor algorithms with instance-weighting have been proposed, for example, Zhu et al. [16] introduced nearest neighbor chain that crosses the classes. This chain progressively links the nearest neighbors belonging to the opposite class and it is used to set the weight similar to the decision plane. Xu et al. [39] proposed weighted rough v-twin support vector machine using *k*-nearest neighbors and Pan et al. [40] developed structural twin support vector machine based on *k*-nearest neighbors. The nearest neighbor concept has also been applied to deal with the class imbalance problem. Ando [41] employed a class-wise weighting with nearest neighbor density estimation to counteract the class imbalance problem and learns its weight parameters by convex optimization. Furthermore, there are some research [13-15] to assign weight through the angle between a neighbor and mean of neighbors, focusing on setting appropriate function of *k*. The sum of cosine values for all neighbors are applied to weighted one-class support vector machine [13], sample reduction [14], and boundary detection [15] for support vector machine.

To further quantify the degree of information's certainty, several papers combining nearest neighbor and entropy have been proposed. Zheng and Zhu [42] proposed intuitionistic fuzzy entropy in a neighborhood system, and Chen et al. [43] developed the information quantity, neighborhood entropy, and information granularity, which express uncertainty in the neighborhood system. Among them, neighborhood entropy is the measure for class certainty level and can be used to formulate the FSVM membership. As an extension of EFSVM, Zhu and Wang [44] developed an entropy-based matrix learning machine using the Matrix-pattern-oriented Ho-Kashyap learning machine (MatMHKS). Also, Gupta et al. [45] combined EFSVM with twin support vector machine.

There have been classification methods without much dependence on the size of nearest neighbors. Pan et al. [18] classified each class by computing the harmonic mean distance of nearest neighbors for each class. Ertugrul and Tagluk [20] measured not only the similarity of nearest neighbors via the distance between two samples, but also the dependency via the angle between samples using adaptive dependency region instead of a fixed size of nearest neighbors. Zhu et al. [21] computed the distance of the fixed radius nearest neighbor using the modified law of gravitation without any manually-set parameters. Zhang et al. [22] proposed dynamic local neighborhood considering the positive-negative border of each instance to adjust the posterior probability of query instances for the minority class.

There have been comparative studies on the effectiveness of each component of IEFSVM for imbalanced classification [38, 46, 47]. We compare the results from IEFSVM with those from canonical SVM (SVM) [48], SVM with undersampling (u-SVM), cost-sensitive SVM (cs-SVM), FSVM [7], and EFSVM [12] by AUC. For the application of SVM methods to imbalanced data classification, u-SVM is a resampling technique belong to preprocessing method [47] and cs-SVM, FSVM [7], EFSVM [12] are cost-sensitive methods.

For the assessment of classifiers, AUC is generally used as an evaluation criterion [49]. Also the performance of IEFSVM is evaluated in comparison with algorithms not based on SVM, AdaBoost [50], Random Forest [51], EasyEnsemble [52], RUSBoost [53], and Weighted-ELM [54]. EasyEnsemble and RUSBoost are ensemble algorithms used as benchmarks in several literatures [55-58]. These two methods combine boosting and random under-sampling techniques, in which EasyEnsemble samples several subsets from the majority class [52], while RUSBoost randomly removes samples from the majority class until the desired balance is achieved [53]. Weighted-ELM generalizes single hidden layer feedforward networks (SLFNs), and each sample is assigned with an extra weight [59]. Weighted-ELM is robust to both balanced and imbalanced data distribution and can be generalized to cost sensitive learning [54]. All of these algorithms include other base classifiers such as decision tree, boosted ensemble and neural network rather than SVM. Therefore, the performance of our algorithm can be compared with state-of-the-art algorithms based on other basic classifiers.

## 3. Method

### 3.1. Entropy fuzzy support vector machine (EFSVM)

#### 3.1.1. Fuzzy support vector machine (FSVM)

Support vector machine (SVM) is the process of deciding the optimal hyperplane with the largest margin, and solves a quadratic optimization problem as follows [48]. Given the training set, $S = \{(x_i, y_i)\}_{i=1}^{N}$, suppose there is a binary-class classification problem with n-dimensional sample $x_i$ and $y_i \in \{-1, 1\}$.

$$min \ \frac{1}{2}w^T w + C \sum_{i=1}^{N} \xi_i$$

$$s.t. \quad y_i(w^T \varphi(x_i) + b) \geq 1 - \xi_i \ , i = 1, \dots, N$$

$$\xi_i \geq 0 \ , i = 1, \dots, N$$

(1)

where $w$ is weight vector of the decision surface, $\varphi(x_i)$ denotes the mapping function into high dimension feature space, $b$ denotes the bias, and $C$ is the regularization parameter to be tuned by the parameter selection.

FSVM is an SVM that weighs each data differently in order to correctly classify some important training samples. If FSVM is applied to the imbalanced data, it can increase the importance of the minority class data by setting a higher weight for the minority class. Unlike SVM, FSVM obtains the hyperplane by multiplying membership ($s_i$) to soft error ($\xi_i$) to adjust the weight. The quadratic optimization problem with training set $S = \{(x_i, y_i, s_i)\}_{i=1}^{N}$ for FSVM is as follows [6].

$$min \ \frac{1}{2}w^T w + C \sum_{i=1}^{N} s_i \xi_i$$

(2)

$$s.t. \quad y_i(w^T\varphi(x_i) + b) \geq 1 - \xi_i \, , i = 1, \ldots, N$$

$$\xi_i \geq 0 \, , i = 1, \ldots, N \, , 0 \leq s_i \leq 1$$

Using the Lagrange multiplier method to solve this optimization problem, equation (2) can be transformed into a dual problem as shown in equation (3) [6].

$$max \sum_{i=1}^{N} \alpha_i - \frac{1}{2}\sum_{i=1}^{N}\sum_{j=1}^{N} \alpha_i \alpha_j y_i y_j K(x_i, x_j) \tag{3}$$

$$s.t. \quad \sum_{i=1}^{N} y_i \alpha_i = 0, \quad 0 \leq \alpha_i \leq s_i C, \quad i = 1, \ldots, N$$

where $K(x_i, x_j) = \varphi(x_i)^T \varphi(x_i)$.

Sequential Minimal Optimization (SMO) [60] is used to solve the above equation, and the optimal values for $\alpha_i$ are achieved. Then, the weight vector and the decision function are as follows [61].

$$w = \sum_{i=1}^{N} \alpha_i y_i \varphi(x_i), \quad f(x) = sign(\sum_{i=1}^{N} \alpha_i y_i K(x, x_i) + b) \tag{4}$$

The difference between FSVM and SVM eventually depends on fuzzy membership, $s_i$. If $s_i$ has a value of one for all $i$, it has the same result as SVM. Thus, when using FSVM, it is important to select appropriate fuzzy membership according to data.

*3.1.2. Entropy fuzzy membership*

The FSVM where the fuzzy membership is determined using entropy of each sample is called entropy fuzzy support vector machine (EFSVM). Entropy of sample $x_i$, $H_i$, is a measure of information certainty [8] using binary classification as follows.

$$H_i = -p_i \ln(p_i) \mathbf{1}\{p_i > 0\} - n_i \ln(n_i) \mathbf{1}\{n_i > 0\} \tag{5}$$

where $p_i$ and $n_i$ are the probabilities that sample $x_i$ belongs to positive and negative class, respectively. Note that $\mathbf{1}\{A\}$ is 1 when $A$ is true and 0 otherwise.

The closer the two probabilities are, the higher the entropy, and vice versa. After searching $k$ nearest neighbors for each sample, we examine which class the nearest neighbors belong to. Let $pos_i$ and $neg_i$ denote the numbers of nearest neighbors belonging to the positive and negative classes, respectively, and $p_i$ and $n_i$ are defined as follows.

$$p_i = pos_i/k, \quad n_i = neg_i/k \tag{6}$$

where $k$ is the number of nearest neighbors.

The entropy for each sample can be obtained from the above equation, and this is called the nearest neighbors entropy. For example, if the entropy is high, it means the information is not clear, so that the fuzzy membership should be low. On the contrary, if entropy is low, it means the information is certain, so that the fuzzy membership should be high. Therefore, there is a negative relationship between the entropy and the membership, and the following equation can be considered as an example of entropy fuzzy membership, $s_i$ [12].

$$s_i = \begin{cases} 1 & \text{when} \quad y_i = +1 \\ (1 - H_i)/IR & \text{when} \quad y_i = -1 \end{cases} \quad (7)$$

There are positive and negative classes as binary classification, and $n_{pos}$ and $n_{neg}$ are the number of positive and negative samples, respectively. The scale of how the number of samples is imbalanced, imbalanced ratio (IR) is defined as $IR = n_{neg}/n_{pos}$. We set a low membership for the majority sample through $IR$ ($IR < 1$). In addition, $(1 - H_i)$ term shows a negative relation between entropy and fuzzy membership. The main processes of EFSVM are outlined in Algorithm 1.

---

**Algorithm 1** EFSVM

**for** $k = 5, 7, 9, 11, 13, 15$ **do**
  **for** $i = 1, ..., |X|$ **do**
    $p_i \leftarrow pos_i/k$
    $n_i \leftarrow neg_i/k$
    $H_i = -p_i \ln(p_i) \mathbf{1}\{p_i > 0\} - n_i \ln(n_i) \mathbf{1}\{n_i > 0\}$
    **if** $y_i = 1$ **then**
      $s_i \leftarrow 1$
    **else if** $y_i = -1$ **then**
      $s_i \leftarrow (1 - H_i)/IR$
    **end if**
  **end for**
  $Error_k \leftarrow$ 5-fold CV error of FSVM with $s_i$
**end for**
$kk \leftarrow \underset{k}{\mathrm{argmin}}\, Error_k$
**for** $i = 1, ..., |X|$ **do**
  $p_i \leftarrow pos_i/kk$
  $n_i \leftarrow neg_i/kk$
  $H_i = -p_i \ln(p_i) \mathbf{1}\{p_i > 0\} - n_i \ln(n_i) \mathbf{1}\{n_i > 0\}$
  **if** $y_i = 1$ **then**
    $ss_i \leftarrow 1$
  **else if** $y_i = -1$ **then**
    $ss_i \leftarrow (1 - H_i)/IR$
  **end if**
**end for**
**return** $ss_i$

---

This entropy fuzzy membership is varied according to the number of nearest neighbors, $k$. This is because, for the whole data, $k$, which guarantees the highest accuracy, is selected. When $k$ is small, we obtain entropies close to the sample and have the risk of overfitting. On the other hand, when $k$ is large, we obtain entropies that cover condition rather away from the sample so that a small amount of information can be ignored and complex distribution is difficult to reflect. Therefore, it is important to set the appropriate number of nearest neighbors for each dataset.

### 3.2. Instance-based entropy fuzzy support vector machine (IEFSVM)

From Eqs. (5) and (6), the factor that changes the entropy is the number of nearest neighbors, if dataset is fixed. The existing EFSVM tunes the number of nearest neighbors, $k$, to maximize the forecasting accuracy. In this paper, we revise the EFSVM in a way that the change of entropy in response to switching $k \in \{1,3,5,7,9,11,13,15\}$ is reflected to the determination of fuzzy membership, $s_i$. The formulation of fuzzy membership is based on the following graphical analysis of nearest neighbors entropy.

*3.2.1. Graphical analysis of nearest neighbors entropy*

For each data point *i*, nearest neighbors entropy can be calculated according to neighborhood size. Let $H_i^k$ be the entropy value of *i* th data point with *k* nearest neighbors. We set the *k* value to (1, 3, 5, 7, 9, 11, 13, 15), and present an example to demonstrate how $H_i^k$ is calculated for each data point *i* in Fig. 1. The squares and triangles in Fig.1 belong to the positive and negative classes, respectively. We can count the number of nearest neighbors belonging to positive class according to the *k* value, which is (0, 0, 0, 0, 0, 1, 3, 5). Then, the corresponding entropy values $\left(H_i^1, H_i^3, H_i^5, H_i^7, H_i^9, H_i^{11}, H_i^{13}, H_i^{15}\right)$ are (0, 0, 0, 0, 0, 0.3046, 0.5402, 0.6365).

[Figure 1 here]

The entropy pairs will vary according to the class ratios of each neighborhood size, and Table 1 shows the description of all nearest neighbors entropy pairs that can be generated by changing the neighborhood size.

[Table 1 here]

In Table 1, the first column counts all different nearest neighbor pairs. The second through ninth columns represent entropy values with each neighborhood size, and the number of nearest neighbors belonging to one class is written in parentheses. For the number of combination of $\{H_i^1, H_i^3, H_i^5, H_i^7, H_i^9, H_i^{11}, H_i^{13}, H_i^{15}\}$, there are 4374 cases. The reason is as follows. When the neighborhood size is one, the number of elements in the positive class can be 0 or 1, which is 2 cases. In other words, the number of combination of $\{H_i^1\}$ is 2. When the neighborhood size is extended to 3 from 1, 3 possible extensions are added to the nearest neighbors: 0,1, or 2 more positive elements. For an example, (1 positive neighbor, 0 negative neighbor) can be extended into the following 3 cases: (3 positive neighbors, 0 negative neighbor), (2 positive neighbors, 1 negative neighbor), (1 positive neighbor, 2 negative neighbors). This implies that the number of combination of $\{H_i^1, H_i^3\}$ is $2 \times 3 = 6$. In this way, the number of combination of $\{H_i^1, H_i^3, H_i^5, \ldots, H_i^{15}\}$ becomes $2 \times 3^7 = 4374$. For each data point, as the neighborhood size increases, the number of nearest neighbors in one class monotonically increases. Specifically for each data point *i*, we consider the average of entropies, $\mu_i = \sum_{k=1}^{8} H_i^{2k-1}/8$, and the standard deviation of entropies, $\sigma_i = \sqrt{\sum_{k=1}^{8}(H_i^{2k-1} - \mu_i)^2/7}$. The tenth and eleventh columns in Table 1 indicate the average and the standard deviation of the entropies, respectively. Then, with all 4374 points, we plot $(\mu_i, \sigma_i)$ in the plane with the horizontal axis of the average and the vertical axis of the standard deviation, as in Figure 2.

[Figure 2 here]

Figure 2 is the scatterplot of the average and the standard deviation of entropies with all 4374 data points in Table 1. As the scatterplot of $(\mu_i, \sigma_i)$ is well fitted in polar coordinate plane, we transform $(\mu_i, \sigma_i)$ into $(d_i, \theta_i)$ where $d_i = \sqrt{\mu_i^2 + \sigma_i^2}$ and $\theta_i = tan^{-1}(\mu_i/\sigma_i)$. Each point, $(d_i, \theta_i)$, has the corresponding entropy information, $\left(H_i^1, H_i^3, H_i^5, H_i^7, H_i^9, H_i^{11}, H_i^{13}, H_i^{15}\right)$, and the points in polar coordinate plane can be classified by the category of the number of nonzero entropies.

Alternatively, the scatterplot appears as a set of lines that point to the origin. These lines need to be analyzed in terms of what they have in common and what is different between the lines, so that we consider the number of nonzero entropies among $H_i^1, H_i^3, \ldots, H_i^{15}$ as a reference. For example in Figure 1, since entropy values $\left(H_i^1, H_i^3, H_i^5, H_i^7, H_i^9, H_i^{11}, H_i^{13}, H_i^{15}\right)$ are (0, 0, 0, 0, 0, 0.3046, 0.5402, 0.6365), there are three nonzero entropies for data point *i*. In this way, data points that have one and two nonzero entropies are specified in Table 2.

[Table 2 here]

Table 2 selects data points from Table 1, which have one and two nonzero nearest neighbors entropies. For $(d_i, \theta_i)$, *i* =2, 3, 4372, 4373, all nearest neighbors whose size up to 13 belong to one class and the neighborhood of different class appears when the nearest neighbor size becomes 15. This results in $H_i^k = 0$ for $k$=1,3,5,7,9,11,13 and $H_i^{15} \neq 0$. For $(d_i, \theta_i)$, $i$ =4, 5, ..., 9, 4366, 4367, ..., 4371, all nearest neighbors whose size up to 11 belong to one class and the neighbors from different class are shown when the nearest neighbor size is over 11. This results in $H_i^k = 0$ for $k$=1,3,5,7,9,11, and $H_i^k \neq 0$ for $k$=13,15. We classify $(d_i, \theta_i)$ based on its number of nonzero entropies, $H_i^k$, in Figure 3. Figure 3 classifies the data points in scatterplot of Figure 2 according to the number of nonzero entropy from one to six. It is observed that $(d_i, \theta_i)$s with the same number of nonzero entropies are linearly located.

[Figure 3 here]

For a closer look at $d_i$ and $\theta_i$ values, let $\theta_i$ be fixed and $d_i$ increase. It means the number of nonzero entropy is fixed, and the data points in the northeast in the scatterplot are selected, so that the average ($\mu_i$) and the standard deviation ($\sigma_i$) of entropies proportionally increase. The increment of both average and standard deviation of entropies of $(d_i, \theta_i)$ implies that the components of the nearest neighbors varies according to the size of nearest neighbors.

The increment of $\theta_i$ implies that the average of entropies $\mu_i$ increases. In Figure 3, it is observed that the number of nonzero entropies for $(d_i, \theta_i)$ and the size of $\theta_i$ has a positive relationship that $(d_i, \theta_i)$ with the more nonzero entropies has the larger $\theta_i$. As the number of nonzero entropies increases, not only $\theta_i$ but also $d_i$ tends to grow as seen in Fig. 3. This is because nonzero entropies increase the values of $\mu_i$ and $\sigma_i$ as seen in Table 1 and Table 2.

### 3.2.2. Instance-based entropy fuzzy membership

Based on the analysis of nearest neighbors entropy, we claim the following four concepts for the determination of $s_i$. First, $s_i$ should be decreased as $\mu_i$ increases as in Eq. (7) because entropy and fuzzy membership has a negative relationship. Second, $s_i$ decreases as $\sigma_i$ increases. High $\sigma_i$ means that there is a large variance between $H_i^1, H_i^3, ..., H_i^{15}$, and the number of nearest neighbors belonging to one class varies for each neighborhood size. Third, $s_i$ should decrease as $\theta_i$ increases. Fourth, $s_i$ should decrease as $d_i$ increases. As high values of $d_i$ and $\theta_i$ are resulted from high values of $\mu_i$ and $\sigma_i$, the points having large values of $d_i$ and $\theta_i$ are not likely to provide a reliable information for classification.

In this polar coordination plane, the given four concepts guarantee that $s_i$ decreases when $d_i$ and $\theta_i$ increase. Based on this idea, we propose a novel method to determine entropy fuzzy membership, $s_i$, using all information from $\{H_i^1, H_i^3, H_i^5, H_i^7, H_i^9, H_i^{11}, H_i^{13}, H_i^{15}\}$.

We propose an instance-based entropy fuzzy membership for IEFSVM considering the information of the nearest neighbors entropy as follows.

$$s_i = \begin{cases} 1 & if \ y_i = +1 \\ \left(1 - (d_i\theta_i - \min_i d_i\theta_i)/(\max_i d_i\theta_i - \min_i d_i\theta_i)\right)/IR & if \ y_i = -1 \end{cases} \quad (8)$$

Note that $\min_i d_i\theta_i$ and $\max_i d_i\theta_i$ are the minimum and the maximum of $d_i\theta_i$ values among all data points, respectively. IEFSVM assigns the fuzzy membership by Eq. (8), and the main procedures of IEFSVM are given in Algorithm 2.

---

**Algorithm 2** IEFSVM
**for** $i = 1, ..., |X|$ **do**
  **for** $k = 1, 3, 5, 7, 9, 11, 13, 15$ **do**
    $p_i \leftarrow pos_i/k$

```
        n_i ← neg_i/k
        H_i^k ← -p_i ln(p_i) 1{p_i > 0} - n_i ln(n_i) 1{n_i > 0}
     end for
     μ_i ← Σ_{k=1}^{8} H_i^{2k-1} / 8
     σ_i ← √(Σ_{k=1}^{8} (H_i^{2k-1} - μ_i)^2 / 7)
     d_i ← √(μ_i^2 + σ_i^2)
     θ_i ← tan^{-1}(μ_i/σ_i)
     if y_i = 1 then
        s_i ← 1
     else if y_i = -1 then
        s_i ← (1 - (d_i θ_i - min_i d_i θ_i)/(max_i d_i θ_i - min_i d_i θ_i))/IR
     end if
  end for
  return s_i
```

In order to visualize IEFSVM, we plot the level curve of proposed fuzzy membership. In Figure 4, the curves corresponding to $d_i \times \theta_i = t$ for $t = 0.15, 0.3, 0.45, 0.6, 0.75$ are added to the scatterplot of $(\mu_i, \sigma_i)$ of all 4374 points.

[Figure 4 here]

## 4. EMPIRICAL STUDY

In this section, we evaluate the performance of IEFSVM in comparison with SVM based methods and five state-of-the-art algorithms on UCI benchmark datasets. Statistical tests are carried out to confirm significant model improvements. SVM based methods consist of canonical SVM [48], SVM with undersampling (u-SVM), cost-sensitive SVM (cs-SVM), FSVM [7], and EFSVM [12], while five other algorithms are cost-sensitive AdaBoost (cs-AdaBoost) [50], cost-sensitive Random Forest (cs-RF) [51], EasyEnsemble [52], RUSBoost [53], and Weighted-ELM (w-ELM) [54].

### 4.1. Experimental settings and datasets

For SVM-based learning machines such as SVM, u-SVM, cs-SVM, FSVM, EFSVM, and IEFSVM, RBF kernel or linear kernel is used according to datasets. The regularization parameter $C$ is chosen from $\{2^{-6}, 2^{-4}, ..., 2^4, 2^6\}$. For tree-based learning machines such as cs-AdaBoost, cs-RF, EasyEnsemble, and RUSBoost, we select 100 maximum learning iterations. For entropy-based fuzzy SVM learning machines such as EFSVM and IEFSVM, the number of nearest neighbors to calculate entropy is chosen from $\{1,3,5,7,9,11,13,15\}$. All of the above parameter tuning procedures are performed through a 5-fold cross validation.

For class imbalance problem, 35 UCI [62] datasets are used to test the performance of our proposed algorithm. IR of the selected datasets ranges from 1.14 to 15.46. Information on this dataset is given in Table 3, which is sorted in ascending order of IR.

[Table 3 here]

In the first column (dataset) of Table 3, the numbers in the 'dataset' title indicate class number. (e.g. sonar1 is the first class of 'sonar' dataset.) The numbers before and next to 'vs' represent the minority and the majority classes, respectively. (e.g. the second and the first classes of wdbc2vs1 are minority and majority classes, respectively.) If there is a number without 'vs', the number is for minority class, and the remaining data belongs to the majority class. (e.g. the first class of sonar1 is minority one and the remaining data are majority.)

The area under the receiver operating characteristics (ROC) curve (AUC) [63] is used for comparison measure of the classification performance of each learning machine. AUC is defined as follows.

$$\text{AUC} = (1 + TP_{rate} - FP_{rate})/2 \tag{9}$$

where, $TP_{rate}$ and $FP_{rate}$ denote the ratio of the positive samples correctly classified and that of the negative samples misclassified, respectively.

## 4.2. Experimental results and analysis

### 4.2.1. Effectiveness of each component of IEFSVM

In order to verify the effectiveness of fuzzy membership used in our model, we conduct experiments to compare the IEFSVM with canonical SVM, u-SVM, cs-SVM, FSVM, EFSVM and IEFSVM. We evaluated the final results by AUC values. Table 4 shows the mean and the standard deviation of AUC values of IEFSVM and the other five methods, and Table 5 shows the AUC rankings of those six algorithms. The mean and the standard deviation of AUC values are calculated from 100 experiments. We set a dataset with IR greater than 3.3 as an imbalanced criterion, so that we specify the average ranking and the average AUC according to the imbalanced criterion such as for all datasets, for datasets with IR greater than 3.3, and for datasets with IR less than 3.3. In Table 4, the best results for each dataset are highlighted in bold. In Table 5, better results between proposed IEFSVM and existing EFSVM for each dataset are highlighted in the shade for comparison.

[Table 4 here]

[Table 5 here]

Overall, the results shown in Table 4 and 5 demonstrate IEFSVM far outperforms other five algorithms when IR is higher than 3.3. While comparing EFSVM with basic SVM algorithms, EFSVM shows slightly better performance. Since EFSVM assigns a greater weight on the minority class, the performance at IR under 3.3 is worse, but the result at IR above 3.3 is a little better than the other basic SVM algorithms. Between EFSVM and IEFSVM, the shaded parts of Table 5 show that EFSVM has a high performance when IR is less than 3.3, but IEFSVM overwhelms EFSVM when IR is higher than 3.3. This is a clear difference, and IEFSVM model is a good fit for imbalanced data with IR higher than 3.3.

Both EFSVM and IEFSVM examine the nearest neighbors in each sample to calculate the entropy. In this respect, there is no difference in computational complexity. Alternatively, EFSVM computes support vectors for all *k* to look for the optimal number of nearest neighbors, while IEFSVM trains a classifier only one time. As we use 8 values {1,3,5,7,9,11,13,15} for *k* in this experiment, the time to determine the support vectors of IEFSVM is eight times shorter than that of EFSVM.

### 4.2.2. Comparison among IEFSVM and other algorithms

In order to check if our proposed IEFSVM outperforms other algorithms using other base classifiers, we conducted experiments with cs-AdaBoost, cs-RF, EasyEnsemble, RUSBoost, and w-ELM. As with previous chapter, we evaluated the final results by AUC values. Table 6 shows the mean and the standard deviation of AUC values of IEFSVM and the other five methods, and Table 7 shows the AUC rankings of those six algorithms. The mean and the standard deviation of AUC values are calculated from 100 repeated experiments. We set an IR greater than 3.3 as an imbalanced criterion and specify the average ranking and the average AUC based on the imbalanced criterion in the category of all datasets, datasets with IR greater than 3.3, datasets with IR less than 3.3. In Table 6, the best results for each dataset are highlighted in bold.

[Table 6 here]

[Table 7 here]

The results show that IEFSVM overwhelms cs-AdaBoost, cs-RF, and RUSBoost, while EasyEnsemble and w-ELM are slightly worse than IEFSVM at IR above 3.3. It implies that IEFSVM model is the best fit for imbalanced problems with IR higher than 3.3. Furthermore, IEFSVM performs better than other algorithms in terms of average AUC value with IR higher than 3.3.

## 4.3. Statistical studies

In order to evaluate the fuzzy membership modification of IEFSVM, we compare performances between IEFSVM and other algorithms. We conduct a Wilcoxon paired signed-rank test [64] for pairwise comparisons. This test is used to compare two algorithms to assess whether their mean ranks differ. To obtain further confirmation that IEFSVM is superior to other algorithms, we perform Holm post hoc test [65], which is a statistical multicomparison test to detect whether several pairs of methods are statistically different. This test is based on the following $z$ value [66].

$$z = (R^* - R)/\sqrt{k(k+1)/(6N)} \tag{10}$$

where $k$ and $N$ denote the number of methods and datasets, respectively. Note that $R^*$ refers to the average ranking of the IEFSVM method for the classification of datasets, and $R$ refers to the average ranking of the other method for the classification of datasets. The $z$ values are sorted in descending order and the corresponding $i$-th method has adjusted alpha for Holm test which equals to $(0.05/i)$. If $p$ value corresponding to the $z$ value is less than the adjusted alpha, the hypothesis that both methods have the same ranking is rejected. In our experiments, $N$ is 35 for all datasets, 23 for the datasets with IR over 3.3 and 12 for the datasets with IR below 3.3.

### 4.3.1. Statistical tests of each component of IEFSVM

The results of Holm post hoc test according to IR value of 3.3 are shown in Table 8. IR below 3.3, IEFSVM is not better than other algorithms, while IEFSVM only gets better performance than SVM for all datasets. IR over 3.3, however, IEFSVM clearly outperforms the other methods with a significant margin. Observing the results of Wilcoxon tests shown in Table 9, the hypothesis test results for IR below 3.3 and for all datasets are the same as Table 8. IR over 3.3, IEFSVM is significantly better than other methods except FSVM. The rank between IEFSVM and FSVM is close and the result of Wilcoxon test does provide any significant evidence of difference between two models.

[Table 8 here]

[Table 9 here]

### 4.3.2. Statistical tests of IEFSVM and other algorithms

The same statistical procedures are performed between IEFSVM and the state-of-the-art algorithms. In Table 10, we tested if there exist significant differences between IEFSVM and other five algorithms using the IR criterion value of 3.3. At IR below 3.3, IEFSVM does not show good performance. However, the hypothesis test results for IR over 3.3 and for all datasets, IEFSVM outperforms cs-AdaBoost, cs-RF, and RUSBoost with significant differences, while IEFSVM is not distinguishable with EasyEnsemble and w-ELM. The results of Wilcoxon tests shown in Table 11 also show the same results as in Holm post hoc test.

[Table 10 here]

[Table 11 here]

## 4.4. The application of IEFSVM algorithms with real-world datasets

In this section, we apply IEFSVM to real-world datasets. The descriptions of the datasets are shown in Section 4.4.1, and the experimental settings are the same as in Section 4.1.

*4.4.1. Data description*

We use 12 real-world datasets to test the practical applicability for class imbalance problem. The data sets used are AIDS [67], Cervical cancer [68], Lending Club [69], Otto group [70], Seoul weather [71, 72]. Since Otto dataset is a multi-class problem, it is transformed into 7 imbalanced binary problems. Information on this dataset is given in Table 12. Numerical variables are normalized in a way that the maximum value is 1 and the minimum value is -1, while categorical variables are binarized to obtain dummy variables. As a preprocess applied to 12 real-world datasets, the gradient boosting method [73] measures the importance of variables and selects variables with high importance.

[Table 12 here]

*4.4.2. Results and analysis*

This section presents the classification results of real-world datasets in terms of AUC with each component of IEFSVM and other state-of-the-art algorithms. Table 13 shows the mean and the standard deviation of AUC values of IEFSVM and the components, and Table 14 shows the AUC rankings of those six algorithms with 100 repeated experiments. Table 14 provides the algorithm rankings from low to high: IEFSVM, EFSVM, FSVM, cs-SVM, u-SVM, and SVM. In addition, the stability of IEFSVM is the best among the SVM based methods although the classification results can be different each time.

[Table 13 here]

[Table 14 here]

Table 15 shows the mean and the standard deviation of AUC values of IEFSVM and five state-of-the-art algorithms, and Table 16 demonstrates the AUC rankings of those six algorithms. According to Table 16, the algorithm rankings from low to high are: IEFSVM, EasyEnsemble, RUSBoost, cs-AdaBoost, w-ELM, and cs-RF. Meanwhile, the stability of IEFSVM is the second best, and EasyEnsemble is the first.

[Table 15 here]

[Table 16 here]

# 5. Conclusion

This paper proposes instance-based entropy fuzzy support vector machine (IEFSVM) to better classify binary imbalanced datasets. In this model, we transform the instance-based entropy into polar coordinate to develop the entropy function that is appropriate for each sample. Considering that existing EFSVM uses a uniform neighborhood size when assigning fuzzy membership, we combine nearest neighbors entropies that change according to neighborhood size for each data point, so that the classifier can assign the fuzzy membership by reflecting all information of each instance efficiently. Another significant contribution of this paper lies in representing graphical analysis of nearest neighbors entropy. This analysis not only explores the pattern of the entropies but also provides rational reasoning to the formula for fuzzy membership.

In experimental studies, IEFSVM is compared with other SVM based methods. The results show that for datasets with IR above 3.3, the proposed algorithm overwhelms other benchmark methods, including existing EFSVM. Furthermore, IEFSVM shows better performance than other state-of-the-art algorithms in the case of IR above 3.3. The fact that IEFSVM works well in UCI and some real-world datasets having IR above 3.3 confirms the superiority of IEFSVM for the classification of real-world imbalanced datasets.

For binary imbalanced data classification, further research on fuzzy membership using nearest neighbors entropy will be followed. An example is the extension of neighborhood size which provides more detail information on the distribution of imbalanced data. In addition, the more elaborate instance-based entropy fuzzy membership than that shown in equation (8) can be developed by the thorough analysis of the polar coordinate distribution of $(d_i, \theta_i)$ which is revised from the neighborhood size extension.

# ACKNOWLEDGEMENTS

This research was supported by the Basic Science Research Program through the National Research Foundation of Korea (NRF) funded by the Ministry of Science, ICT and Future Planning (NRF-2015R1A2A2A03005488).

**Table 1** Description of all nearest neighbors entropies according to neighborhood size

| $i$ | $H_i^1$ | $H_i^3$ | $H_i^5$ | $H_i^7$ | $H_i^9$ | $H_i^{11}$ | $H_i^{13}$ | $H_i^{15}$ | $\mu_i$ | $\sigma_i$ |
|---|---|---|---|---|---|---|---|---|---|---|
| 1 | 0(0) | 0(0) | 0(0) | 0(0) | 0(0) | 0(0) | 0(0) | 0(0) | 0 | 0 |
| 2 | 0(0) | 0(0) | 0(0) | 0(0) | 0(0) | 0(0) | 0(0) | 0.2449(1) | 0.0306 | 0.0866 |
| 3 | 0(0) | 0(0) | 0(0) | 0(0) | 0(0) | 0(0) | 0(0) | 0.3927(2) | 0.0491 | 0.1388 |
| 4 | 0(0) | 0(0) | 0(0) | 0(0) | 0(0) | 0(0) | 0.2712(1) | 0.2449(1) | 0.0645 | 0.1197 |
| 5 | 0(0) | 0(0) | 0(0) | 0(0) | 0(0) | 0(0) | 0.2712(1) | 0.3927(2) | 0.0830 | 0.1570 |
| 6 | 0(0) | 0(0) | 0(0) | 0(0) | 0(0) | 0(0) | 0.2712(1) | 0.5004(3) | 0.0964 | 0.1888 |
| 7 | 0(0) | 0(0) | 0(0) | 0(0) | 0(0) | 0(0) | 0.4293(2) | 0.3927(2) | 0.1027 | 0.1905 |
| 8 | 0(0) | 0(0) | 0(0) | 0(0) | 0(0) | 0(0) | 0.4293(2) | 0.5004(3) | 0.1162 | 0.2160 |
| 9 | 0(0) | 0(0) | 0(0) | 0(0) | 0(0) | 0(0) | 0.4293(2) | 0.5799(4) | 0.1262 | 0.2370 |
| 10 | 0(0) | 0(0) | 0(0) | 0(0) | 0(0) | 0.3046(1) | 0.2712(1) | 0.2449(1) | 0.1026 | 0.1425 |
| 11 | 0(0) | 0(0) | 0(0) | 0(0) | 0(0) | 0.3046(1) | 0.2712(1) | 0.3927(2) | 0.1211 | 0.1704 |
| - | - | - | - | - | - | - | - | - | - | - |
| 4374 | 0(1) | 0(3) | 0(5) | 0(7) | 0(9) | 0(11) | 0(13) | 0(15) | 0 | 0 |

**Table 2** Description of nearest neighbors entropies with one and two nonzero values.

| $i$ | $H_i^1$ | $H_i^3$ | $H_i^5$ | $H_i^7$ | $H_i^9$ | $H_i^{11}$ | $H_i^{13}$ | $H_i^{15}$ | $\mu_i$ | $\sigma_i$ |
|---|---|---|---|---|---|---|---|---|---|---|
| 2 | 0(0) | 0(0) | 0(0) | 0(0) | 0(0) | 0(0) | 0(0) | **0.2449(1)** | 0.0306 | 0.0866 |
| 3 | 0(0) | 0(0) | 0(0) | 0(0) | 0(0) | 0(0) | 0(0) | **0.3927(2)** | 0.0491 | 0.1388 |
| 4372 | 0(1) | 0(3) | 0(5) | 0(7) | 0(9) | 0(11) | 0(13) | **0.3927(13)** | 0.0491 | 0.1388 |
| 4373 | 0(1) | 0(3) | 0(5) | 0(7) | 0(9) | 0(11) | 0(13) | **0.2449(14)** | 0.0306 | 0.0866 |
| 4 | 0(0) | 0(0) | 0(0) | 0(0) | 0(0) | 0(0) | **0.2712(1)** | **0.2449(1)** | 0.0645 | 0.1197 |
| 5 | 0(0) | 0(0) | 0(0) | 0(0) | 0(0) | 0(0) | **0.2712(1)** | **0.3927(2)** | 0.0830 | 0.1570 |
| 6 | 0(0) | 0(0) | 0(0) | 0(0) | 0(0) | 0(0) | **0.2712(1)** | **0.5004(3)** | 0.0964 | 0.1888 |
| 7 | 0(0) | 0(0) | 0(0) | 0(0) | 0(0) | 0(0) | **0.4293(2)** | **0.3927(2)** | 0.1027 | 0.1905 |
| 8 | 0(0) | 0(0) | 0(0) | 0(0) | 0(0) | 0(0) | **0.4293(2)** | **0.5004(3)** | 0.1162 | 0.2160 |
| 9 | 0(0) | 0(0) | 0(0) | 0(0) | 0(0) | 0(0) | **0.4293(2)** | **0.5799(4)** | 0.1262 | 0.2370 |
| 4366 | 0(1) | 0(3) | 0(5) | 0(7) | 0(9) | 0(11) | **0.4293(11)** | **0.5799(11)** | 0.1262 | 0.2370 |
| 4367 | 0(1) | 0(3) | 0(5) | 0(7) | 0(9) | 0(11) | **0.4293(11)** | **0.5004(12)** | 0.1162 | 0.2160 |
| 4368 | 0(1) | 0(3) | 0(5) | 0(7) | 0(9) | 0(11) | **0.4293(11)** | **0.3927(13)** | 0.1027 | 0.1905 |
| 4369 | 0(1) | 0(3) | 0(5) | 0(7) | 0(9) | 0(11) | **0.2712(12)** | **0.5004(12)** | 0.0964 | 0.1888 |
| 4370 | 0(1) | 0(3) | 0(5) | 0(7) | 0(9) | 0(11) | **0.2712(12)** | **0.3927(13)** | 0.0830 | 0.1570 |
| 4371 | 0(1) | 0(3) | 0(5) | 0(7) | 0(9) | 0(11) | **0.2712(12)** | **0.2449(14)** | 0.0645 | 0.1197 |

**Table 3** Information of imbalanced UCI datasets

| dataset | IR | Inst. | Pos. | Neg. | Dim | dataset | IR | Inst. | Pos. | Neg. | Dim |
|---|---|---|---|---|---|---|---|---|---|---|---|
| sonar1 | 1.14 | 208 | 97 | 111 | 60 | ecoli178vs46 | 9.09 | 222 | 22 | 200 | 7 |
| liver2vs1 | 1.38 | 345 | 145 | 200 | 6 | ecoli1345vs6 | 9.1 | 202 | 20 | 182 | 7 |
| wdbc2vs1 | 1.68 | 569 | 212 | 357 | 30 | ecoli157vs6 | 9.15 | 203 | 20 | 183 | 7 |
| ionosphere0 | 1.79 | 351 | 126 | 225 | 33 | ecoli12vs346 | 9.17 | 244 | 24 | 220 | 7 |
| glass2 | 1.82 | 214 | 76 | 138 | 9 | ecoli1378vs46 | 9.18 | 224 | 22 | 202 | 7 |
| diabetes1 | 1.87 | 768 | 268 | 500 | 8 | glass15vs6 | 9.22 | 92 | 9 | 83 | 9 |
| iris2 | 2 | 150 | 50 | 100 | 4 | ecoli1457vs6 | 9.25 | 205 | 20 | 185 | 7 |
| seeds1 | 2 | 210 | 70 | 140 | 7 | zoo1257vs6 | 9.38 | 83 | 8 | 75 | 16 |
| glass1 | 2.06 | 214 | 70 | 144 | 9 | ecoli178vs6 | 10 | 220 | 20 | 200 | 7 |
| dermatology456 | 2.09 | 358 | 116 | 242 | 34 | abalone9vs16 | 10.28 | 756 | 67 | 689 | 8 |
| haberman1vs2 | 2.78 | 306 | 81 | 225 | 3 | ecoli12vs6 | 11 | 240 | 20 | 220 | 7 |
| transfusion1 | 3.2 | 748 | 178 | 570 | 4 | abalone10vs4 | 11.12 | 691 | 57 | 634 | 8 |
| ecoli2 | 3.36 | 336 | 77 | 259 | 7 | zoo | 11.63 | 101 | 8 | 93 | 16 |
| heart0 | 3.85 | 267 | 55 | 212 | 44 | abalone9vs17 | 11.88 | 747 | 58 | 689 | 8 |
| ecoli8 | 5.46 | 336 | 52 | 284 | 7 | abalone9vs4 | 12.09 | 746 | 57 | 689 | 8 |
| dermatology4 | 6.46 | 358 | 48 | 310 | 34 | ecoli1257vs6 | 13 | 280 | 20 | 260 | 7 |
| dermatology5 | 6.46 | 358 | 48 | 310 | 34 | glass5 | 15.46 | 214 | 13 | 201 | 9 |
| ecoli5 | 8.6 | 336 | 35 | 301 | 7 | | | | | | |

**Table 4** Comparison of AUC obtained via each component of IEFSVM on UCI datasets (Note that: the best results are highlighted in **BOLD**).

| Dataset | IR | SVM | u-SVM | cs-SVM | Fuzzy SVM | EFSVM | IEFSVM |
|---|---|---|---|---|---|---|---|
| sonar1 | 1.14 | **75.83+-4** | 75.69+-4.12 | 74.89+-4.53 | 75.57+-3.86 | 74.83+-3.89 | 73.64+-4.32 |
| liver2vs1 | 1.38 | 67.64+-3.19 | 66.32+-4.03 | 67.14+-4.02 | **68.36+-3.42** | 60.74+-3.31 | 56.27+-3.13 |
| wdbc2vs1 | 1.68 | 96.27+-1.35 | 96.58+-1.29 | 96.43+-1.21 | 96.46+-1.27 | **96.74+-1.19** | 96.1+-1.58 |
| ionosphere0 | 1.79 | 93.72+-1.76 | 92+-2.49 | **93.82+-1.8** | 93.69+-1.73 | 93.78+-1.83 | 93.8+-1.82 |
| glass2 | 1.82 | 71.49+-3.93 | 71.62+-4.32 | **73.29+-4.38** | 72.94+-3.77 | 72.56+-4.58 | 72.47+-3.85 |
| diabetes1 | 1.87 | 71.24+-2.72 | 72.96+-2.07 | 73.61+-2.2 | **74.01+-2.18** | 72.95+-2.37 | 72.06+-2.02 |
| iris2 | 2 | **95.05+-2.57** | 94.1+-2.68 | 94.57+-3.03 | 94.93+-2.92 | 94.73+-2.59 | 94.18+-2.8 |
| seeds1 | 2 | **91.91+-2.93** | 91.8+-2.9 | 91.79+-2.49 | 91.59+-2.19 | 91.68+-2.57 | 91.38+-2.36 |
| glass1 | 2.06 | 79.42+-5.7 | 78.95+-5.8 | **81.38+-4.98** | 81.27+-5.49 | 81.36+-5.35 | 78.88+-4.42 |
| dermatology456 | 2.09 | **95.63+-1.77** | 94.46+-2.26 | 95.49+-1.57 | 95.35+-1.95 | 94.99+-1.94 | 94.89+-1.51 |
| haberman1vs2 | 2.78 | 56.2+-3.67 | 61.65+-4.6 | **62.31+-4.2** | 61+-3.95 | 59.25+-4.67 | 59.24+-4.7 |
| transfusion1 | 3.2 | 60.5+-3.71 | **67.23+-2.5** | 66.88+-2.72 | 66.89+-2.59 | 65.03+-2.52 | 63.49+-2.43 |
| ecoli2 | 3.36 | 85.32+-3.55 | 88.5+-2.91 | 88.7+-2.55 | 88.26+-2.17 | 89.07+-2.47 | **89.2+-2.26** |
| heart0 | 3.85 | 64.04+-5.05 | 73.89+-3.52 | 74.56+-3.65 | 75.07+-4 | 74.87+-3.71 | **76.29+-3.36** |
| ecoli8 | 5.46 | 91.73+-3.47 | 91.01+-3.7 | 92.82+-2.35 | 92.82+-2.65 | 92.86+-2.59 | **93.1+-2.59** |
| dermatology4 | 6.46 | 93.49+-3.13 | 92.58+-2.73 | 94.7+-2.82 | 95.04+-2.54 | 95.03+-2.81 | **95.57+-2.17** |
| dermatology5 | 6.46 | 98.96+-2.3 | 99.5+-0.97 | 99.76+-0.68 | **99.78+-0.54** | 99.49+-1.19 | 99.73+-0.67 |
| ecoli5 | 8.6 | 79.74+-6.57 | 87.14+-3.52 | **87.84+-3.47** | 87.77+-3.82 | 87.62+-3.85 | 87.77+-3.4 |
| ecoli178vs46 | 9.09 | 84.62+-6.51 | 89.14+-5.12 | 87.74+-5.19 | **89.76+-5.39** | 89.16+-5.04 | 89.17+-5.05 |
| ecoli1345vs6 | 9.1 | 94.44+-4.4 | 96.75+-2.61 | 96.55+-3.23 | 96.04+-3.6 | 96.35+-3.16 | **97.2+-2.7** |
| ecoli157vs6 | 9.15 | 93.02+-6.12 | **95.99+-2.9** | 94.98+-3.42 | 94.1+-3.48 | 94.71+-3.34 | 94.97+-2.9 |
| ecoli12vs346 | 9.17 | 89.5+-6.34 | 91.97+-4.28 | **92.91+-4.83** | 92.56+-5.47 | 92.38+-5.05 | 92.47+-4.32 |
| ecoli1378vs46 | 9.18 | 85.46+-6.17 | 87.7+-4.94 | 87.51+-4.85 | 86.95+-5.53 | 87.75+-5.24 | **88.05+-4.7** |
| glass15vs6 | 9.22 | 83.67+-10.97 | 83.32+-11.4 | 84.52+-10.5 | 87.39+-10.5 | 86.14+-10.79 | **89.83+-8.63** |
| ecoli1457vs6 | 9.25 | 94.42+-5.19 | **95.68+-3.92** | 94.65+-3.87 | 94.89+-3.66 | 94.95+-3.65 | 95.48+-3.71 |
| zoo1257vs6 | 9.38 | 96.7+-7.33 | 90.63+-8.23 | 96.47+-6.53 | 96.73+-7.77 | 97.03+-5.72 | **97.27+-7.03** |
| ecoli178vs6 | 10 | 90.11+-6.02 | **92.62+-3.83** | 91.23+-4.4 | 90.51+-5.15 | 92.39+-4.86 | 92.49+-4.13 |
| abalone9vs16 | 10.28 | 74.63+-3.5 | 82.42+-4.67 | 83.7+-3.8 | 83.93+-3.59 | **84.1+-3.29** | 83.86+-4.21 |
| ecoli12vs6 | 11 | 94.92+-3.94 | **97.27+-2.83** | 95.5+-3.4 | 95.7+-3.71 | 96.3+-3.81 | 96.2+-3.05 |
| abalone10vs4 | 11.12 | 97.55+-2.38 | 97.41+-1.98 | 97.59+-1.08 | **97.68+-1.01** | 97.45+-1.3 | 97.55+-0.84 |
| zoo6 | 11.63 | 94.35+-12.02 | 93.78+-5.09 | 95.7+-8.97 | 95.5+-9.78 | 96.7+-7.77 | **98.51+-1.34** |
| abalone9vs17 | 11.88 | 71.39+-4.55 | 82.28+-4.91 | **84.36+-4.2** | 83.71+-4.11 | 84.01+-3.53 | 84.17+-3.84 |
| abalone9vs4 | 12.09 | 96.7+-2.3 | **97.67+-1.16** | 97.52+-1.37 | 97.59+-1.21 | 97.31+-1.36 | 97.45+-1.01 |
| ecoli1257vs6 | 13 | 93.54+-5.3 | **96.22+-3.42** | 93.26+-4.98 | 92.93+-5.41 | 94.19+-3.68 | 95+-4.29 |
| glass5 | 15.46 | 77.26+-9.76 | 83.86+-9.83 | 85.24+-8.75 | 87.46+-7.38 | **89.33+-5.79** | 87.08+-9.17 |
| Average AUC | | 85.16+-4.69 | 87.16+-3.93 | 87.7+-3.89 | **87.84+-3.94** | 87.65+-3.74 | 87.57+-3.44 |
| IR below 3.3 | | 79.58+-3.11 | 80.28+-3.26 | 80.97+-3.09 | **81.01+-2.94** | 79.89+-3.07 | 78.87+-2.91 |
| IR over 3.3 | | 88.07+-5.52 | 90.75+-4.28 | 91.21+-4.3 | 91.4+-4.46 | 91.7+-4.09 | **92.1+-3.71** |

**Table 5** AUC rankings of the each component of IEFSVM on UCI datasets (Note that: better results are highlighted by shading between IEFSVM and EFSVM).

| Dataset | IR | SVM | u-SVM | cs-SVM | Fuzzy SVM | EFSVM | IEFSVM |
|---|---|---|---|---|---|---|---|
| sonar1 | 1.14 | 1 | 2 | 4 | 3 | 5 | 6 |
| liver2vs1 | 1.38 | 2 | 4 | 3 | 1 | 5 | 6 |
| wdbc2vs1 | 1.68 | 5 | 2 | 4 | 3 | 1 | 6 |
| ionosphere0 | 1.79 | 4 | 6 | 1 | 5 | 3 | 2 |
| glass2 | 1.82 | 6 | 5 | 1 | 2 | 3 | 4 |
| diabetes1 | 1.87 | 6 | 3 | 2 | 1 | 4 | 5 |
| iris2 | 2 | 1 | 6 | 4 | 2 | 3 | 5 |
| seeds1 | 2 | 1 | 2 | 3 | 5 | 4 | 6 |
| glass1 | 2.06 | 4 | 5 | 1 | 3 | 2 | 6 |
| dermatology456 | 2.09 | 1 | 6 | 2 | 3 | 4 | 5 |
| haberman1vs2 | 2.78 | 6 | 2 | 1 | 3 | 4 | 5 |
| transfusion1 | 3.2 | 6 | 1 | 3 | 2 | 4 | 5 |
| ecoli2 | 3.36 | 6 | 4 | 3 | 5 | 2 | 1 |
| heart0 | 3.85 | 6 | 5 | 4 | 2 | 3 | 1 |
| ecoli8 | 5.46 | 5 | 6 | 4 | 3 | 2 | 1 |
| dermatology4 | 6.46 | 5 | 6 | 4 | 2 | 3 | 1 |
| dermatology5 | 6.46 | 6 | 4 | 2 | 1 | 5 | 3 |
| ecoli5 | 8.6 | 6 | 5 | 1 | 3 | 4 | 2 |
| ecoli178vs46 | 9.09 | 6 | 4 | 5 | 1 | 3 | 2 |
| ecoli1345vs6 | 9.1 | 6 | 2 | 3 | 5 | 4 | 1 |
| ecoli157vs6 | 9.15 | 6 | 1 | 2 | 5 | 4 | 3 |
| ecoli12vs346 | 9.17 | 6 | 5 | 1 | 2 | 4 | 3 |
| ecoli1378vs46 | 9.18 | 6 | 3 | 4 | 5 | 2 | 1 |
| glass15vs6 | 9.22 | 5 | 6 | 4 | 2 | 3 | 1 |
| ecoli1457vs6 | 9.25 | 6 | 1 | 5 | 4 | 3 | 2 |
| zoo1257vs6 | 9.38 | 4 | 6 | 5 | 3 | 2 | 1 |
| ecoli178vs6 | 10 | 6 | 1 | 4 | 5 | 3 | 2 |
| abalone9vs16 | 10.28 | 6 | 5 | 4 | 2 | 1 | 3 |
| ecoli12vs6 | 11 | 6 | 1 | 5 | 4 | 2 | 3 |
| abalone10vs4 | 11.12 | 3 | 6 | 2 | 1 | 5 | 4 |
| zoo6 | 11.63 | 5 | 6 | 3 | 4 | 2 | 1 |
| abalone9vs17 | 11.88 | 6 | 5 | 1 | 4 | 3 | 2 |
| abalone9vs4 | 12.09 | 6 | 1 | 3 | 2 | 5 | 4 |
| ecoli1257vs6 | 13 | 4 | 1 | 5 | 6 | 3 | 2 |
| glass5 | 15.46 | 6 | 5 | 4 | 2 | 1 | 3 |
| Average rank | | 4.86 | 3.8 | 3.06 | 3.03 | 3.17 | 3.09 |
| IR below 3.3 | | 3.58 | 3.67 | 2.42 | 2.75 | 3.5 | 5.08 |
| IR over 3.3 | | 5.52 | 3.87 | 3.39 | 3.17 | 3 | 2.04 |

**Table 6** Comparison of AUC between IEFSVM and five state-of-the-art algorithms on UCI datasets (Note that: the best results are highlighted in **BOLD**).

| Dataset | IR | cs-AdaBoost | cs-RF | EasyEnsemble | RUSBoost | w-ELM | IEFSVM |
|---|---|---|---|---|---|---|---|
| sonar1 | 1.14 | 79.18+-3.55 | 79.01+-4.99 | 81.31+-4.21 | 72.15+-3.75 | **84.8+-4.37** | 73.64+-4.32 |
| liver2vs1 | 1.38 | 68.32+-4.31 | 67.27+-3.45 | **69.21+-3.25** | 58.19+-4.6 | 60.35+-3.81 | 56.27+-3.13 |
| wdbc2vs1 | 1.68 | 94.93+-1.48 | 94.75+-1.47 | 95.74+-1.6 | 91+-1.8 | 94.08+-1.38 | **96.1+-1.58** |
| ionosphere0 | 1.79 | 88.96+-2.72 | 91.57+-1.8 | 91.89+-1.9 | 78.7+-5.67 | 83.89+-2.48 | **93.8+-1.82** |
| glass2 | 1.82 | 72.05+-4.94 | 78.36+-4.57 | **79.74+-4.86** | 66.05+-4.67 | 69.02+-4.02 | 72.47+-3.85 |
| diabetes1 | 1.87 | 73.06+-2.56 | 72.32+-2.57 | **73.49+-2.11** | 70.35+-3.23 | 70.55+-2.05 | 72.06+-2.02 |
| iris2 | 2 | 94+-2.51 | **94.22+-2.55** | 93.8+-2.86 | 94.02+-2.76 | 89.18+-3.58 | 94.18+-2.8 |
| seeds1 | 2 | **91.98+-2.67** | 90.46+-2.85 | 90.59+-2.81 | 87.71+-4.39 | 90.04+-2.53 | 91.38+-2.36 |
| glass1 | 2.06 | 80.31+-4.84 | **85.59+-4.94** | 85.15+-4.07 | 80.44+-4.17 | 70.62+-2.6 | 78.88+-4.42 |
| dermatology456 | 2.09 | 94.84+-1.71 | 95.94+-1.23 | 95.97+-1.69 | 83.61+-10.64 | **96.07+-1.33** | 94.89+-1.51 |
| haberman1vs2 | 2.78 | **63.95+-4.21** | 59.52+-4.31 | 62.28+-3.8 | 63.89+-3.99 | 59.57+-4.23 | 59.24+-4.7 |
| transfusion1 | 3.2 | 67.8+-2.37 | 62.83+-2.73 | 64.18+-2.3 | **67.86+-2.31** | 67.23+-2.4 | 63.49+-2.43 |
| ecoli2 | 3.36 | 88.15+-3.23 | 87.99+-2.92 | 89.52+-2.21 | **89.87+-2.44** | 87.61+-2.37 | 89.2+-2.26 |
| heart0 | 3.85 | 71.68+-4.56 | 68.63+-4.44 | 75.12+-3.99 | 73.95+-4.28 | 70.57+-2.42 | **76.29+-3.36** |
| ecoli8 | 5.46 | 88.35+-3.35 | 88.66+-4.31 | 89.3+-2.89 | 86.84+-3.47 | 89.42+-2.84 | **93.1+-2.59** |
| dermatology4 | 6.46 | 93.36+-3.16 | 93.83+-3.32 | 94.23+-2.68 | 88.39+-3.55 | 94.3+-1.23 | **95.57+-2.17** |
| dermatology5 | 6.46 | 90.85+-19.33 | 99.82+-0.64 | 74.08+-24.39 | 90.82+-19.32 | **99.85+-0.28** | 99.73+-0.67 |
| ecoli5 | 8.6 | 84.11+-7.21 | 79.4+-6.98 | 87.11+-3.94 | 87.03+-4.27 | 87.06+-2.32 | **87.77+-3.4** |
| ecoli178vs46 | 9.09 | 86.24+-6.37 | 85.83+-5.83 | 88.24+-4.51 | 87.92+-5.21 | **91.68+-4.23** | 89.17+-5.05 |
| ecoli1345vs6 | 9.1 | 94.77+-6.09 | 92.85+-5.86 | 96.81+-2.76 | 95.29+-4.66 | **98.11+-1.87** | 97.2+-2.7 |
| ecoli157vs6 | 9.15 | 93.6+-6.46 | 92.46+-7.65 | 95.03+-3.43 | 93.48+-6.14 | **95.95+-2.52** | 94.97+-2.9 |
| ecoli12vs346 | 9.17 | 87.99+-5.58 | 86.95+-7.18 | 88.9+-4.33 | 87.82+-5.26 | **95.62+-3.03** | 92.47+-4.32 |
| ecoli1378vs46 | 9.18 | 85.28+-5.84 | 86.15+-5.64 | 86.53+-5.72 | 86.95+-4.89 | **89.69+-5.67** | 88.05+-4.7 |
| glass15vs6 | 9.22 | 77.51+-24.72 | **98.42+-3.84** | 92.14+-6.67 | 94.96+-8.16 | 92.49+-5.82 | 89.83+-8.63 |
| ecoli1457vs6 | 9.25 | 94.4+-5.74 | 91.61+-7.76 | 95.75+-4.1 | 92.23+-5.6 | **96.58+-2.36** | 95.48+-3.71 |
| zoo1257vs6 | 9.38 | 92.9+-11.58 | 97.63+-5.31 | 87.47+-8.57 | 95.23+-5.23 | **97.83+-4.69** | 97.27+-7.03 |
| ecoli178vs6 | 10 | 91.72+-5.21 | 91.11+-6.35 | **94.04+-4.33** | 93.19+-4.27 | 92.25+-3.15 | 92.49+-4.13 |
| abalone9vs16 | 10.28 | 77.25+-3.82 | 72.09+-4.08 | 79.32+-3.22 | 72.94+-3.46 | 68.2+-5.09 | **83.86+-4.21** |
| ecoli12vs6 | 11 | 91.18+-6.55 | 90.43+-7.91 | 97.63+-1.69 | 93.28+-6.86 | **98.7+-0.75** | 96.2+-3.05 |
| abalone10vs4 | 11.12 | 84.66+-21.89 | **98.14+-1.53** | 98.12+-1.23 | 95.35+-11.64 | 97.62+-1.05 | 97.55+-0.84 |
| zoo6 | 11.63 | 93.3+-11.71 | 98.2+-4.76 | 87.05+-10.37 | 95.28+-7.98 | 96.51+-7.82 | **98.51+-1.34** |
| abalone9vs17 | 11.88 | 78.35+-4.81 | 72.1+-4.3 | 83.04+-3.69 | 68.83+-4.23 | 67.4+-6.11 | **84.17+-3.84** |
| abalone9vs4 | 12.09 | 91.38+-15.58 | **97.9+-1.96** | 97.49+-1.42 | 97.46+-2.12 | 97.47+-1.06 | 97.45+-1.01 |
| ecoli1257vs6 | 13 | 91.56+-7.82 | 89.85+-8.95 | **96.19+-2.84** | 93.55+-6.67 | 95.9+-2.4 | 95+-4.29 |
| glass5 | 15.46 | 80.18+-11.75 | 83.73+-10.62 | **88.15+-5.28** | 87.73+-7.35 | 86.16+-6.2 | 87.08+-9.17 |
| Average AUC | | 85.09+-6.86 | 86.16+-4.56 | 86.99+-4.28 | 84.64+-5.4 | 86.35+-3.15 | **87.57+-3.44** |
| IR below 3.3 | | 80.78+-3.16 | 80.99+-3.12 | **81.94+-2.95** | 76.16+-4.33 | 77.95+-2.9 | 78.87+-2.91 |
| IR over 3.3 | | 87.34+-8.8 | 88.86+-5.31 | 89.62+-4.97 | 89.06+-5.96 | 90.74+-3.27 | **92.1+-3.71** |

**Table 7** AUC rankings between IEFSVM and five state-of-the-art algorithms on UCI datasets.

| Dataset | IR | cs-AdaBoost | cs-RF | EasyEnsemble | RUSBoost | w-ELM | IEFSVM |
|---|---|---|---|---|---|---|---|
| sonar1 | 1.14 | 3 | 4 | 2 | 6 | 1 | 5 |
| liver2vs1 | 1.38 | 2 | 3 | 1 | 5 | 4 | 6 |
| wdbc2vs1 | 1.68 | 3 | 4 | 2 | 6 | 5 | 1 |
| ionosphere0 | 1.79 | 4 | 3 | 2 | 6 | 5 | 1 |
| glass2 | 1.82 | 4 | 2 | 1 | 6 | 5 | 3 |
| diabetes1 | 1.87 | 2 | 3 | 1 | 6 | 5 | 4 |
| iris2 | 2 | 4 | 1 | 5 | 3 | 6 | 2 |
| seeds1 | 2 | 1 | 4 | 3 | 6 | 5 | 2 |
| glass1 | 2.06 | 4 | 1 | 2 | 3 | 6 | 5 |
| dermatology456 | 2.09 | 5 | 3 | 2 | 6 | 1 | 4 |
| haberman1vs2 | 2.78 | 1 | 5 | 3 | 2 | 4 | 6 |
| transfusion1 | 3.2 | 2 | 6 | 4 | 1 | 3 | 5 |
| ecoli2 | 3.36 | 4 | 5 | 2 | 1 | 6 | 3 |
| heart0 | 3.85 | 4 | 6 | 2 | 3 | 5 | 1 |
| ecoli8 | 5.46 | 5 | 4 | 3 | 6 | 2 | 1 |
| dermatology4 | 6.46 | 5 | 4 | 3 | 6 | 2 | 1 |
| dermatology5 | 6.46 | 4 | 2 | 6 | 5 | 1 | 3 |
| ecoli5 | 8.6 | 5 | 6 | 2 | 4 | 3 | 1 |
| ecoli178vs46 | 9.09 | 5 | 6 | 3 | 4 | 1 | 2 |
| ecoli1345vs6 | 9.1 | 5 | 6 | 3 | 4 | 1 | 2 |
| ecoli157vs6 | 9.15 | 4 | 6 | 2 | 5 | 1 | 3 |
| ecoli12vs346 | 9.17 | 4 | 6 | 3 | 5 | 1 | 2 |
| ecoli1378vs46 | 9.18 | 6 | 5 | 4 | 3 | 1 | 2 |
| glass15vs6 | 9.22 | 6 | 1 | 4 | 2 | 3 | 5 |
| ecoli1457vs6 | 9.25 | 4 | 6 | 2 | 5 | 1 | 3 |
| zoo1257vs6 | 9.38 | 5 | 2 | 6 | 4 | 1 | 3 |
| ecoli178vs6 | 10 | 5 | 6 | 1 | 2 | 4 | 3 |
| abalone9vs16 | 10.28 | 3 | 5 | 2 | 4 | 6 | 1 |
| ecoli12vs6 | 11 | 5 | 6 | 2 | 4 | 1 | 3 |
| abalone10vs4 | 11.12 | 6 | 1 | 2 | 5 | 3 | 4 |
| zoo6 | 11.63 | 5 | 2 | 6 | 4 | 3 | 1 |
| abalone9vs17 | 11.88 | 3 | 4 | 2 | 5 | 6 | 1 |
| abalone9vs4 | 12.09 | 6 | 1 | 2 | 4 | 3 | 5 |
| ecoli1257vs6 | 13 | 5 | 6 | 1 | 4 | 2 | 3 |
| glass5 | 15.46 | 6 | 5 | 1 | 2 | 4 | 3 |
| Average rank | | 4.14 | 4 | 2.63 | 4.2 | 3.17 | 2.86 |
| IR below 3.3 | | 2.92 | 3.25 | 2.33 | 4.67 | 4.17 | 3.67 |
| IR over 3.3 | | 4.78 | 4.39 | 2.78 | 3.96 | 2.65 | 2.43 |

**Table 8** Holm test results of the each component of IEFSVM.

| All datasets | | | | | IR below 3.3 | | | | | IR over 3.3 | | | | |
|---|---|---|---|---|---|---|---|---|---|---|---|---|---|---|
| Algorithms | Z | p-Value | Holm | Hypothesis | Algorithms | Z | p-Value | Holm | Hypothesis | Algorithms | Z | p-Value | Holm | Hypothesis |
| SVM | 4.8571 | 0 | 0.01 | Rejected | u-SVM | 3.6667 | 0.9682 | 0.01 | Not rejected | SVM | 5.5217 | 0 | 0.01 | Rejected |
| u-SVM | 3.8 | 0.0551 | 0.0125 | Not rejected | SVM | 3.5833 | 0.9752 | 0.0125 | Not rejected | u-SVM | 3.8696 | 0.0005 | 0.0125 | Rejected |
| EFSVM | 3.1714 | 0.424 | 0.0167 | Not rejected | EFSVM | 3.5 | 0.9809 | 0.0167 | Not rejected | cs-SVM | 3.3913 | 0.0073 | 0.0167 | Rejected |
| cs-SVM | 3.0571 | 0.5255 | 0.025 | Not rejected | FSVM | 2.75 | 0.9989 | 0.025 | Not rejected | FSVM | 3.1739 | 0.0202 | 0.025 | Rejected |
| FSVM | 3.0286 | 0.5508 | 0.05 | Not rejected | cs-SVM | 2.4167 | 0.9998 | 0.05 | Not rejected | EFSVM | 3 | 0.0415 | 0.05 | Rejected |

**Table 9** Wilcoxon test results of the each component of IEFSVM.

| All datasets | | | | IR below 3.3 | | | | IR over 3.3 | | | |
|---|---|---|---|---|---|---|---|---|---|---|---|
| Algorithm | Z | p-Value | Hypothesis(α=0.05) | Algorithm | Z | p-Value | Hypothesis(α=0.05) | Algorithm | Z | p-Value | Hypothesis(α=0.05) |
| SVM | -2.7299 | 0.0063 | Rejected | cs-SVM | 3.6578 | 0.0003 | Not rejected | SVM | -3.6849 | 0.0002 | Rejected |
| u-SVM | -1.3677 | 0.1714 | Not rejected | FSVM | 2.0509 | 0.0403 | Not rejected | cs-SVM | -2.0668 | 0.0388 | Rejected |
| EFSVM | -0.6461 | 0.5182 | Not rejected | EFSVM | 1.6895 | 0.0911 | Not rejected | u-SVM | -2.0265 | 0.0427 | Rejected |
| cs-SVM | -0.149 | 0.8815 | Not rejected | u-SVM | 0.8085 | 0.4188 | Not rejected | EFSVM | -1.9841 | 0.0472 | Rejected |
| FSVM | 0.1156 | 0.908 | Not rejected | SVM | 0.5174 | 0.6049 | Not rejected | FSVM | -1.2127 | 0.2253 | Not rejected |

**Table 10** Holm test results of other state-of-the-art algorithms.

| All datasets | | | | | IR below 3.3 | | | | | IR over 3.3 | | | | |
|---|---|---|---|---|---|---|---|---|---|---|---|---|---|---|
| Algorithms | Z | p-Value | Holm | Hypothesis | Algorithms | Z | p-Value | Holm | Hypothesis | Algorithms | Z | p-Value | Holm | Hypothesis |
| RUSBoost | 4.2 | 0.0013 | 0.01 | Rejected | RUSBoost | 4.6667 | 0.0952 | 0.01 | Not rejected | cs-AdaBoost | 4.7826 | 0 | 0.01 | Rejected |
| cs-AdaBoost | 4.1429 | 0.002 | 0.0125 | Rejected | w-ELM | 4.1667 | 0.2563 | 0.0125 | Not rejected | cs-RF | 4.3913 | 0.0002 | 0.0125 | Rejected |
| cs-RF | 4 | 0.0053 | 0.0167 | Rejected | cs-RF | 3.25 | 0.7073 | 0.0167 | Not rejected | RUSBoost | 3.9565 | 0.0029 | 0.0167 | Rejected |
| w-ELM | 3.1714 | 0.2411 | 0.025 | Not rejected | cs-AdaBoost | 2.9167 | 0.8369 | 0.025 | Not rejected | EasyEnsemble | 2.7826 | 0.2642 | 0.025 | Not rejected |
| EasyEnsemble | 2.6286 | 0.6954 | 0.05 | Not rejected | EasyEnsemble | 2.3333 | 0.9596 | 0.05 | Not rejected | w-ELM | 2.6522 | 0.3468 | 0.05 | Not rejected |

**Table 11** Wilcoxon test results of other state-of-the-art algorithms.

| All datasets | | | | IR below 3.3 | | | | IR over 3.3 | | | |
|---|---|---|---|---|---|---|---|---|---|---|---|
| Algorithm | Z | p-Value | Hypothesis(α=0.05) | Algorithm | Z | p-Value | Hypothesis(α=0.05) | Algorithm | Z | p-Value | Hypothesis(α=0.05) |
| cs-AdaBoost | -2.9578 | 0.0031 | Rejected | RUSBoost | -1.9528 | 0.0508 | Not rejected | cs-AdaBoost | -3.4097 | 0.0007 | Rejected |
| RUSBoost | -2.8795 | 0.004 | Rejected | w-ELM | -1.8533 | 0.0638 | Not rejected | RUSBoost | -2.6543 | 0.0079 | Rejected |
| cs-RF | -2.3133 | 0.0207 | Rejected | EasyEnsemble | 1.0464 | 0.2954 | Not rejected | cs-RF | -2.4185 | 0.0156 | Rejected |
| EasyEnsemble | 0.7316 | 0.4644 | Not rejected | cs-RF | -0.4333 | 0.6648 | Not rejected | w-ELM | 0.8206 | 0.4119 | Not rejected |
| w-ELM | -0.4639 | 0.6427 | Not rejected | cs-AdaBoost | -0.3755 | 0.7073 | Not rejected | EasyEnsemble | 0.3562 | 0.7217 | Not rejected |

Table 12 Information of imbalanced real-world datasets.

| dataset | IR | Inst. | Pos. | Neg. | Dim. |
|---|---|---|---|---|---|
| AIDS | 26.15 | 38529 | 1419 | 37110 | 6 |
| Cancer | 13.84 | 668 | 45 | 623 | 31 |
| Lending | 6.63 | 331879 | 43480 | 288399 | 32 |
| Otto1 | 31.08 | 61878 | 1929 | 59949 | 93 |
| Otto3 | 6.73 | 61878 | 8004 | 53874 | 93 |
| Otto4 | 21.99 | 61878 | 2691 | 59187 | 93 |
| Otto5 | 21.59 | 61878 | 2739 | 59139 | 93 |
| Otto7 | 20.8 | 61878 | 2839 | 59039 | 93 |
| Otto8 | 6.31 | 61878 | 8464 | 53414 | 93 |
| Otto9 | 11.49 | 61878 | 4955 | 56923 | 93 |
| weather1 | 5.31 | 7677 | 1216 | 6461 | 8 |
| weather2 | 16.91 | 12880 | 719 | 12161 | 6 |

Table 13 Comparison of AUC obtained via each component of IEFSVM on real-world datasets (Note that: the best results are highlighted in **BOLD**).

| | IR | SVM | u-SVM | cs-SVM | FSVM | EFSVM | IEFSVM |
|---|---|---|---|---|---|---|---|
| AIDS | 26.15 | 75.09+-11.13 | 85.89+-9.47 | 85.84+-8.92 | 84.72+-9.78 | **86.03+-10.08** | 85.84+-9.72 |
| Cancer | 13.84 | 82.15+-10.14 | 90.78+-5.24 | **91.91+-3.55** | 91.67+-3.64 | 91.45+-4.17 | 91.77+-3.84 |
| Lending | 6.63 | 51.39+-2.01 | 57.19+-4.98 | 59.77+-4.12 | 58.21+-5.39 | 57.24+-5.47 | **60.85+-2.42** |
| Otto1 | 31.08 | 59.28+-9.51 | 69.53+-9.35 | 71.08+-10.6 | 73.16+-10.9 | 72.69+-9.63 | **73.37+-9.61** |
| Otto3 | 6.73 | 60.57+-4.68 | 74.11+-5.39 | 75.94+-5.14 | 75.95+-4.74 | 76.42+-4.15 | **77.15+-4.21** |
| Otto4 | 21.99 | 53.38+-4.97 | 66.38+-8.03 | 67.11+-8.72 | 67.01+-9.66 | 69.76+-8.81 | **70.78+-8.09** |
| Otto5 | 21.59 | 90.95+-6.79 | 92.75+-4.9 | 93.58+-5.38 | **95.57+-3.73** | 94.8+-5.33 | 94.9+-5.15 |
| Otto7 | 20.80 | 65.9+-10.57 | 67.81+-8.19 | 72.52+-8.4 | 73.6+-9.01 | 73.49+-8.8 | **74.43+-8.2** |
| Otto8 | 6.31 | 82.52+-6.43 | 85.77+-4.16 | 88.17+-4.27 | 88.31+-4.18 | **88.78+-4.41** | 88.72+-3.73 |
| Otto9 | 11.49 | 79.92+-7.21 | 82.96+-6.56 | 85.88+-6.98 | 87.02+-6.11 | 87.24+-5.98 | **87.48+-5.84** |
| weather1 | 5.31 | 50+-0 | 66.42+-2.09 | 67.05+-1.82 | 67.82+-2.26 | 66.95+-2.88 | **70.01+-2.68** |
| weather2 | 16.91 | 62.22+-7.79 | 76.02+-3.48 | 75.99+-4.87 | 75.28+-3.5 | 75.79+-5.77 | **76.81+-4.95** |
| Average | | 67.78+-13.27 | 76.3+-10.76 | 77.9+-10.49 | 78.19+-10.79 | 78.39+-10.82 | **79.34+-9.86** |

Table 14 AUC rankings of the each component of IEFSVM on real-world datasets (Note that: better results are highlighted by shading between IEFSVM and EFSVM).

| | IR | SVM | u-SVM | cs-SVM | FSVM | EFSVM | IEFSVM |
|---|---|---|---|---|---|---|---|
| AIDS | 26.15 | 6 | 2 | 3 | 5 | 1 | 4 |
| Cancer | 13.84 | 6 | 5 | 1 | 3 | 4 | 2 |
| Lending | 6.63 | 6 | 5 | 2 | 3 | 4 | 1 |
| Otto1 | 31.08 | 6 | 5 | 4 | 2 | 3 | 1 |
| Otto3 | 6.73 | 6 | 5 | 4 | 3 | 2 | 1 |
| Otto4 | 21.99 | 6 | 5 | 3 | 4 | 2 | 1 |
| Otto5 | 21.59 | 6 | 5 | 4 | 1 | 3 | 2 |
| Otto7 | 20.8 | 6 | 5 | 4 | 2 | 3 | 1 |
| Otto8 | 6.31 | 6 | 5 | 4 | 3 | 1 | 2 |
| Otto9 | 11.49 | 6 | 5 | 4 | 3 | 2 | 1 |
| weather1 | 5.31 | 6 | 5 | 3 | 2 | 4 | 1 |
| weather2 | 16.91 | 6 | 2 | 3 | 5 | 4 | 1 |
| Average rank | | 6 | 4.5 | 3.25 | 3 | 2.75 | 1.5 |

**Table 15** Comparison of AUC between IEFSVM and five state-of-the-art algorithms on real-world datasets (Note that: the best results are highlighted in **BOLD**).

|          | IR    | cs-AdaBoost   | cs-RF         | EasyEnsemble  | RUSBoost      | w-ELM         | IEFSVM        |
|----------|-------|---------------|---------------|---------------|---------------|---------------|---------------|
| AIDS     | 26.15 | 87.68+-7.74   | 85.34+-9.51   | 90.48+-5.6    | **90.7+-6.55**| 75.94+-9.81   | 85.84+-9.72   |
| Cancer   | 13.84 | 91.05+-4.39   | 85.25+-7.04   | 90.73+-3.14   | **91.91+-3.55**| 85.54+-4.77  | 91.77+-3.84   |
| Lending  | 6.63  | 57.86+-3.84   | 52.37+-2.34   | 55.31+-7.35   | 58.6+-3.05    | 53.72+-3.01   | **60.85+-2.42**|
| Otto1    | 31.08 | 63.04+-9.78   | 52.78+-5.49   | 68.99+-10.96  | 67.08+-9.53   | 64.4+-7.55    | **73.37+-9.61**|
| Otto3    | 6.73  | 71.51+-5.63   | 64.24+-5.79   | 76.45+-4.4    | 68.63+-6.18   | 74.45+-6.34   | **77.15+-4.21**|
| Otto4    | 21.99 | 64.32+-7.62   | 54.01+-5.3    | **74.39+-7.85**| 69.3+-8.29   | 71.19+-7.12   | 70.78+-8.09   |
| Otto5    | 21.59 | 92.26+-8.16   | 86.04+-8.67   | 93.77+-7.46   | 94.18+-5.38   | 72.33+-7.91   | **94.9+-5.15**|
| Otto7    | 20.80 | 69.06+-8.99   | 55.8+-7       | 72.38+-8.9    | **74.45+-8.03**| 69.41+-8.88  | 74.43+-8.2    |
| Otto8    | 6.31  | 83.23+-3.88   | 82.36+-4.19   | 88.59+-3.32   | 74.89+-5.17   | 79.09+-7.54   | **88.72+-3.73**|
| Otto9    | 11.49 | 79.04+-5.95   | 72.14+-6.58   | 83.94+-5.21   | 79.01+-7.39   | 81.7+-7.66    | **87.48+-5.84**|
| weather1 | 5.31  | 62+-3.85      | 69.95+-2.61   | 69.51+-2.27   | 61.24+-3.62   | 66.51+-2.02   | **70.01+-2.68**|
| weather2 | 16.91 | **78.06+-4.96**| 65.93+-3.07  | 73.09+-7.74   | 76.08+-4.5    | 76.92+-4.61   | 76.81+-4.95   |
| Average  |       | 74.93+-11.46  | 68.85+-12.85  | 78.14+-11     | 75.51+-11.21  | 72.6+-8.16    | **79.34+-9.86**|

**Table 16** AUC rankings between IEFSVM and five state-of-the-art algorithms on real-world datasets.

|              | IR    | cs-AdaBoost | cs-RF | EasyEnsemble | RUSBoost | w-ELM | IEFSVM |
|--------------|-------|-------------|-------|--------------|----------|-------|--------|
| AIDS         | 26.15 | 3           | 5     | 2            | 1        | 6     | 4      |
| Cancer       | 13.84 | 3           | 6     | 4            | 1        | 5     | 2      |
| Lending      | 6.63  | 3           | 6     | 4            | 2        | 5     | 1      |
| Otto1        | 31.08 | 5           | 6     | 2            | 3        | 4     | 1      |
| Otto3        | 6.73  | 4           | 6     | 2            | 5        | 3     | 1      |
| Otto4        | 21.99 | 5           | 6     | 1            | 4        | 2     | 3      |
| Otto5        | 21.59 | 4           | 5     | 3            | 2        | 6     | 1      |
| Otto7        | 20.80 | 5           | 6     | 3            | 1        | 4     | 2      |
| Otto8        | 6.31  | 3           | 4     | 2            | 6        | 5     | 1      |
| Otto9        | 11.49 | 4           | 6     | 2            | 5        | 3     | 1      |
| weather1     | 5.31  | 5           | 2     | 3            | 6        | 4     | 1      |
| weather2     | 16.91 | 1           | 6     | 5            | 4        | 2     | 3      |
| Average rank |       | 3.75        | 5.33  | 2.75         | 3.33     | 4.08  | 1.75   |

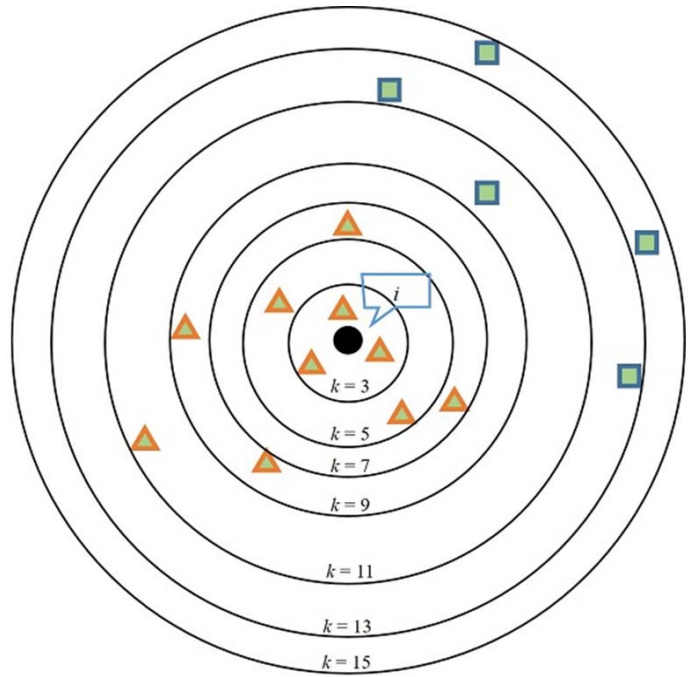

**Fig. 1** Demonstration on nearest neighbors for a data point.

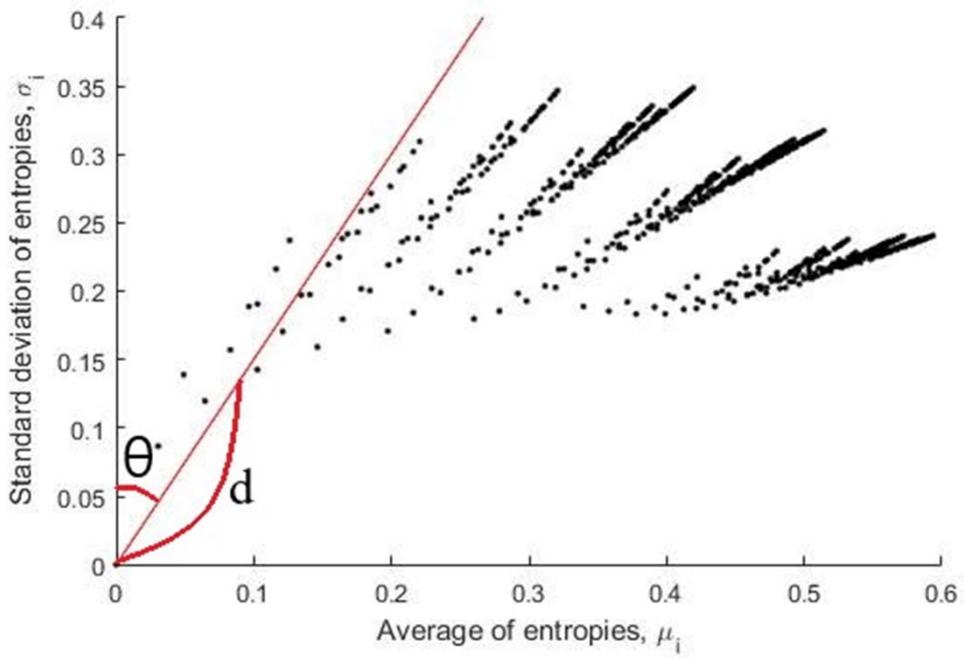

**Fig. 2** Scatterplot of average and standard deviation of all entropies.

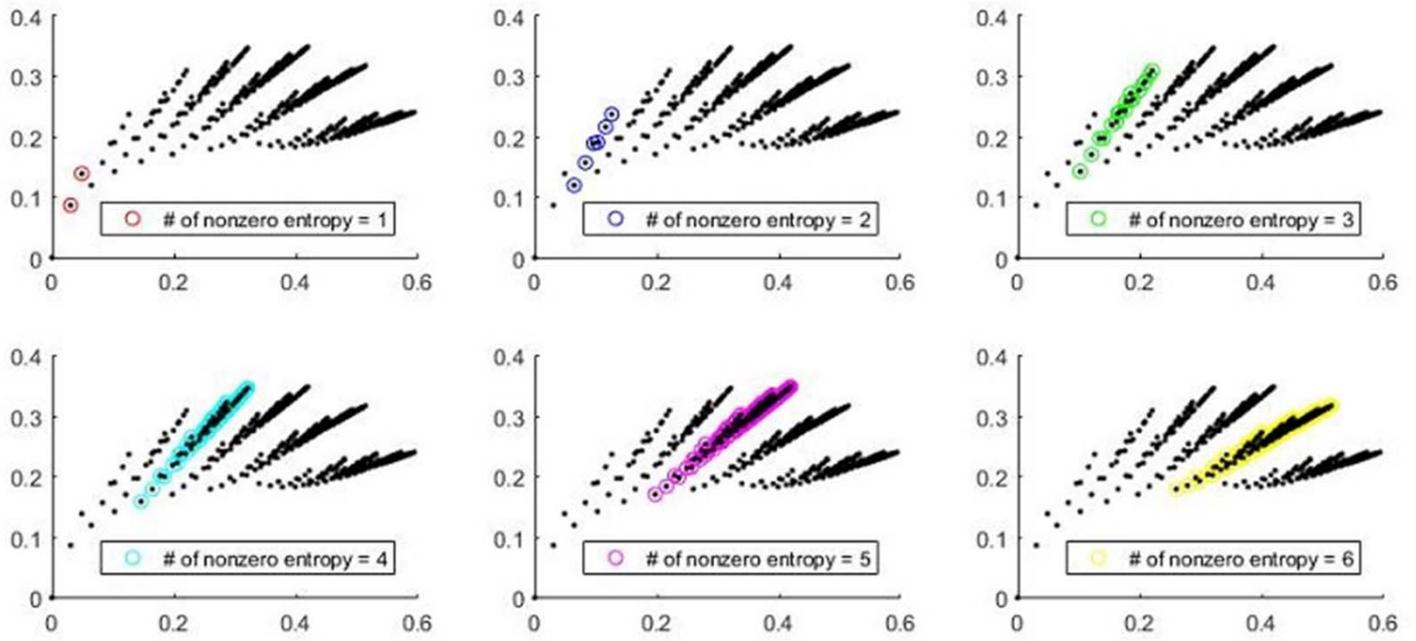

**Fig. 3** Scatterplots according to the number of nonzero nearest neighbors entropies.

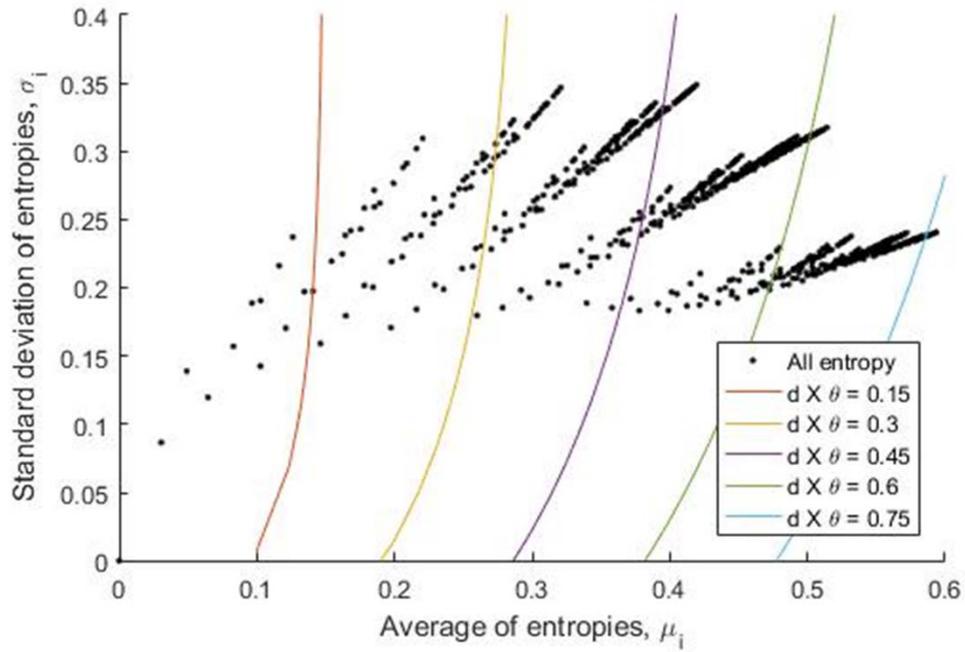

**Fig. 4** Scatterplot of all entropy and level curve of proposed fuzzy membership.